\documentclass{article} 
\usepackage[preprint]{colm2025_conference}

\usepackage{microtype}
\usepackage{hyperref}
\usepackage{url}
\usepackage{booktabs}
\usepackage{graphicx}
\usepackage{colortbl}
\usepackage{tabularx} 
\usepackage{array}
\usepackage{lineno}
\usepackage[toc,page]{appendix}
\usepackage{titletoc}
\usepackage{adjustbox}
\usepackage{makecell}
\usepackage{multicol}
\usepackage{multirow}
\usepackage{algorithm}
\usepackage[noend]{algpseudocode}

\definecolor{darkblue}{rgb}{0, 0, 0.5}
\hypersetup{colorlinks=true, citecolor=darkblue, linkcolor=darkblue, urlcolor=darkblue}

\definecolor{ngreen}{HTML}{76B900}

\usepackage{amsmath,amsfonts,bm}
\usepackage{xspace}
\usepackage{fancyvrb}
\usepackage{fvextra} 
\usepackage[most]{tcolorbox}
\usepackage{footmisc}
\usepackage{eso-pic}
\usepackage{graphicx}
\usepackage{transparent}

\newcommand{\ouralgo}{Retro-Search\xspace}

\newcommand{\model}{\mathcal{M}}           
\newcommand{\question}{q}                  
\newcommand{\questionspace}{\mathcal{Q}}   
\newcommand{\answer}{a}                    
\newcommand{\answerspace}{\mathcal{A}}     

\newcommand{\reasoning}{T}                 
\newcommand{\reasoningspace}{\mathcal{T}}  
\newcommand{\betterreasoning}{\reasoning^*}
\newcommand{\thought}[1]{s^{#1}}           

\newcommand{\stepinthought}[2]{s^{#1}_{#2}} 

\newcommand\BackgroundPic{
    \put(-10,295){
        \parbox[b][\paperheight]{\paperwidth}{
            \vfill
            \centering
            \transparent{0.18}
            \includegraphics[width=1.1\paperwidth,height=0.4\paperheight]{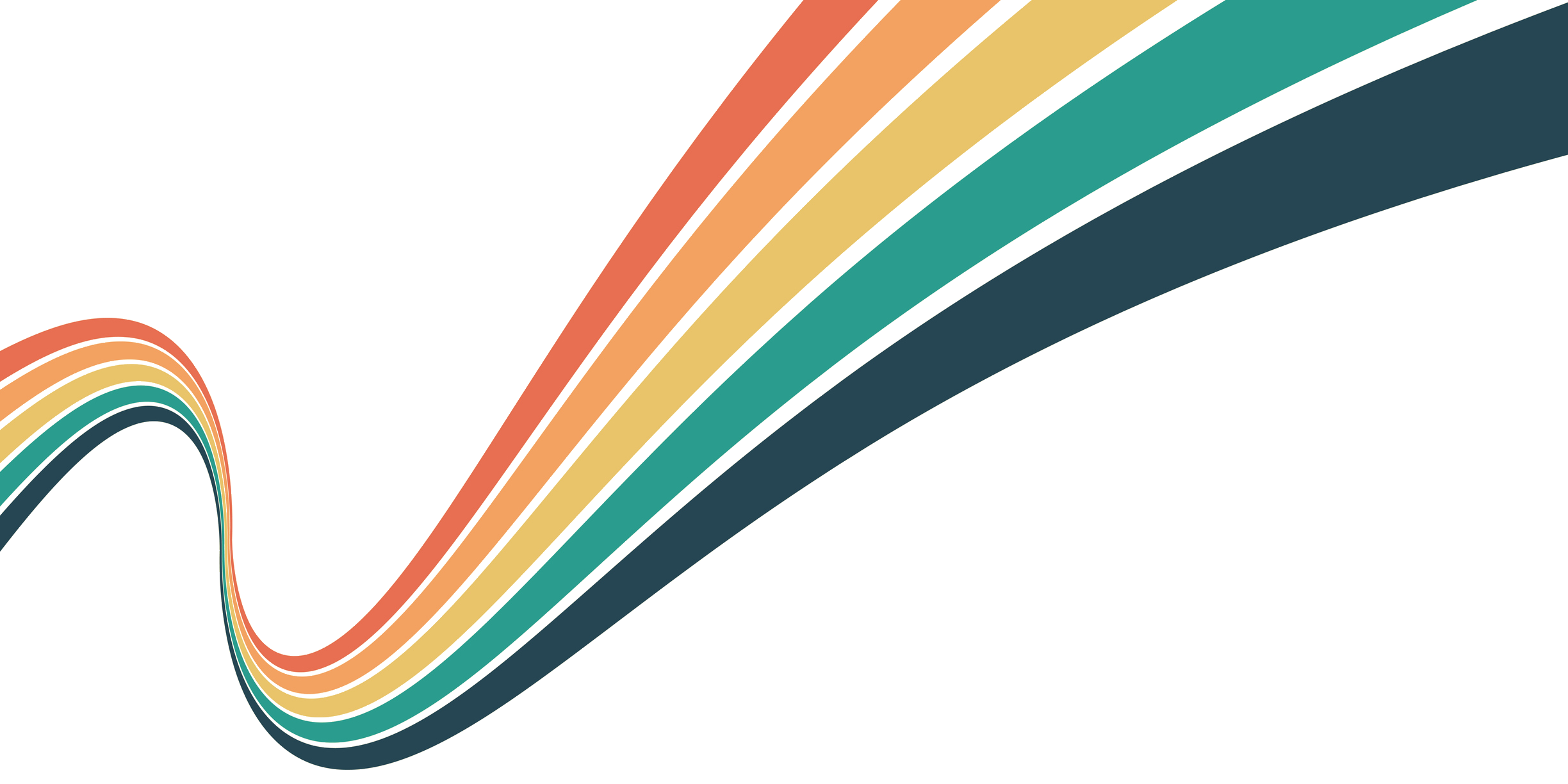}
            \vfill
        }
    }
}

\lstdefinestyle{mystyle}{
    columns=fullflexible,
    backgroundcolor=\color{backcolour},   
    commentstyle=\color{codegreen},
    numberstyle=\tiny\color{codegray},
    stringstyle=\color{codepurple},
    basicstyle=\footnotesize,
    breakatwhitespace=false,         
    breaklines=true,  
    breakindent=0pt,
    captionpos=b,                    
    keepspaces=true,  
    numbersep=0pt,                  
    showspaces=false,                
    showstringspaces=false,
    showtabs=true,                  
    tabsize=10,
    escapeinside={@}{@}, 
    moredelim=**[is][\highlight]{`}{`}, 
}
\title{\textit{Retro-Search}:\\Exploring Untaken Paths for Deeper and Efficient Reasoning}

\author{Ximing Lu$^\dagger$$^\ddagger$\textcolor{ngreen}{$^\clubsuit$}\hspace{0.17cm}
Seungju Han$^\dagger$$^\S$\textcolor{ngreen}{$^\clubsuit$}\hspace{0.17cm}
David Acuna$^\dagger$\textcolor{ngreen}{$^\clubsuit$}\hspace{0.17cm}Hyunwoo Kim$^\dagger$\textcolor{ngreen}{$^\clubsuit$}\hspace{0.17cm}Jaehun Jung$^\dagger$$^\ddagger$\textcolor{ngreen}{$^\clubsuit$}\\
\textbf{Shrimai Prabhumoye$^\dagger$\hspace{1.cm}Niklas Muennighoff\hspace{0.03cm}$^\S$\hspace{1.cm}Mostofa Patwary$^\dagger$}\\
\textbf{Mohammad Shoeybi$^\dagger$\hspace{1.cm}
Bryan Catanzaro$^\dagger$\hspace{1.cm}Yejin Choi$^\dagger$} \\
[0.1cm]
$^\dagger$NVIDIA\hspace{1.1cm}$^\ddagger$University of Washington\hspace{1.1cm}$^\S$Stanford University \\
\texttt{\{ximingl,\hspace{0.05cm}seungjuh,\hspace{0.05cm}dacunamarrer,\hspace{0.05cm}hyunwook,\hspace{0.05cm}jaehunj,\hspace{0.05cm}yejin\}\hspace{0.05cm}@nvidia.com}
}

\begin{document}
\ifcolmsubmission
\linenumbers
\fi

\maketitle
\AddToShipoutPicture*{\BackgroundPic}

\begingroup
\renewcommand\thefootnote{\textcolor{ngreen}{$\clubsuit$}}
\footnotetext{\hspace{0.1cm} First co-authors.}
\endgroup

\begin{abstract}

Large reasoning models, such as OpenAI o1 and DeepSeek-R1, demonstrate remarkable reasoning capabilities via long, elaborate reasoning trajectories. 
Numerous follow-up studies report that supervised fine-tuning on such reasoning traces, also known as distillation, can be a cost-effective way to boost reasoning capabilities of smaller student models.  
However, empirical observations reveal that these reasoning trajectories are often suboptimal, switching excessively between different lines of thought, resulting in under-thinking, over-thinking, and even degenerate responses.
%
%
In this work, we introduce \textit{\ouralgo}, a search algorithm in the spirit of Monte-Carlo Tree Search, for distilling higher quality reasoning paths from large reasoning models. \textit{\ouralgo} retrospectively revises reasoning paths to discover better, yet shorter traces, which can then lead to student models with enhanced reasoning capabilities with shorter, thus faster inference. 
Our approach can enable two use cases: \textbf{self-improvement}, where models are fine-tuned on their own \textit{\ouralgo}-ed thought traces, and \textbf{weak-to-strong improvement}, where a weaker model revises stronger model's thought traces via \textit{\ouralgo}. 
For self-improving, R1-distill-7B, fine-tuned on its own \textit{\ouralgo}-ed traces, reduces the average reasoning length by 31.2\% while improving performance by 7.7\% across seven math benchmarks.
For weak-to-strong improvement, we retrospectively revise R1-671B's traces from the OpenThoughts dataset \citep{openthoughts} using R1-distill-32B as the \textit{\ouralgo}-er, a model 20× smaller.
Qwen2.5-32B, fine-tuned on 40k instances of this refined data, achieves performance comparable to R1-distill-32B, yielding an 11.3\% reduction in reasoning length and a 2.4\% performance improvement compared to fine-tuning on the original OpenThoughts data. 
More excitingly, R1-distill-7B and R1-distill-32B, fine-tuned on this revised data, achieve new state-of-the-art reasoning performance at the 7B and 32B scales while yielding the highest inference efficiency.
%
%
%
Our work counters recently emergent viewpoints that question the relevance of search algorithms in the era of large reasoning models, by demonstrating that there are still opportunities for algorithmic advancements, even for frontier models. 




\end{abstract}

\begin{figure}[t!]
    \centering
    \includegraphics[width=0.95\textwidth]{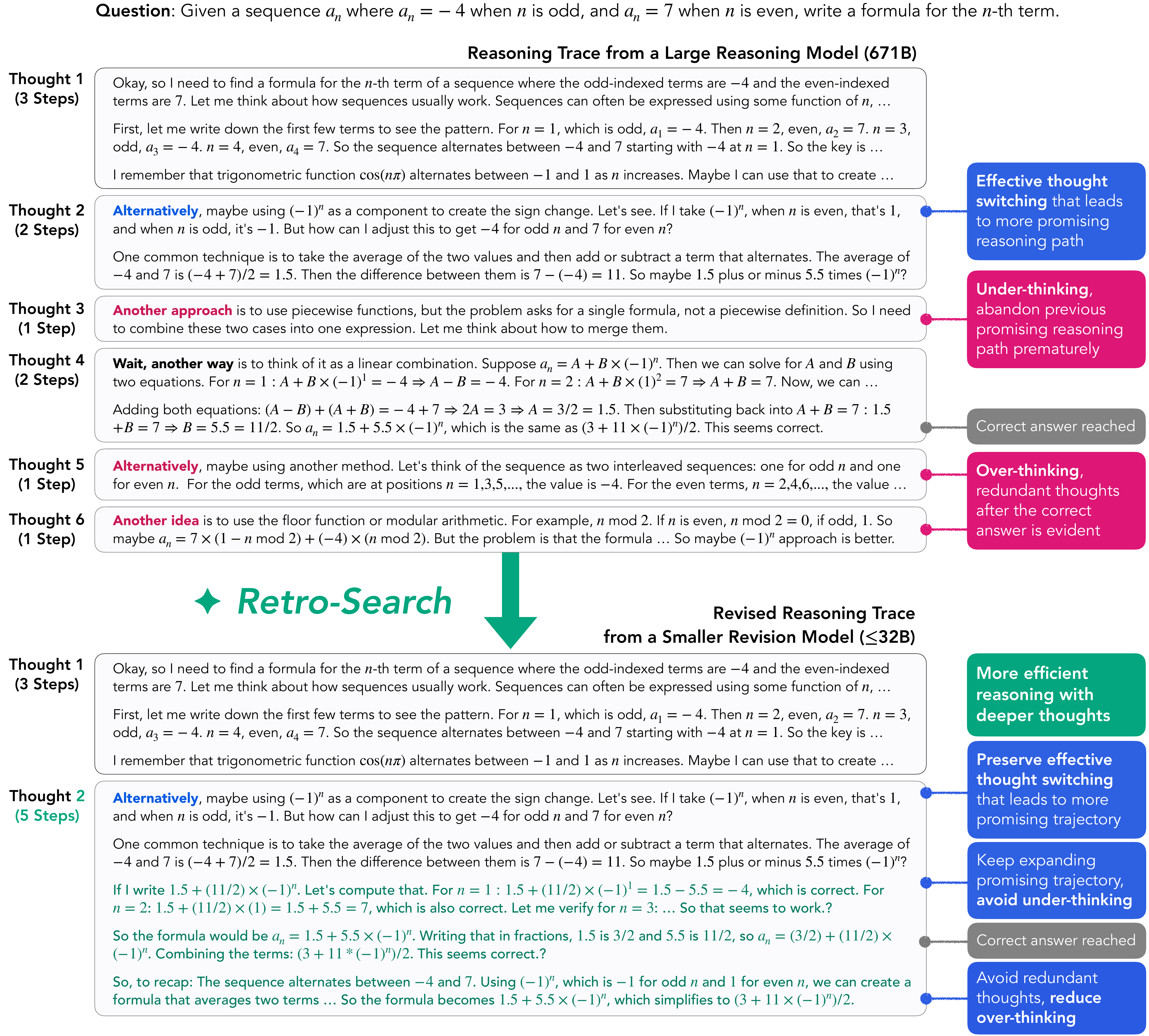}
    \vspace{-5pt}
    \caption{
        An example reasoning trace from \textit{\ouralgo} in weak-to-strong revision. 
A reasoning trace consists of a series of thoughts segmented by transition keywords (e.g., “alternatively”, “wait”), with each thought composed of a sequence of intermediate steps, delimited by '\texttt{\textbackslash n\textbackslash n}'.
\textit{\ouralgo} retrospectively revises reasoning trajectories - 
exploring promising thoughts that were prematurely abandoned to mitigate under-thinking while avoiding redundant thoughts once the correct answer is evident to reduce over-thinking.
        %
    }
    \vspace{-5pt}
    \label{fig:enter-label}
\end{figure}

\section{Introduction}\label{sec:intro}

Recent state-of-the-art LLMs, such as OpenAI o1 and DeepSeek-R1, have demonstrated remarkable capabilities in solving complex reasoning problems by scaling test-time compute. Test-time scaling enables the model to produce extended reasoning trajectories---an inner monologue akin to an implicit internal search---where the model explores multiple potential solution paths and verifies itself~\citep{o1, deepseekai2025deepseekr1incentivizingreasoningcapability, qwq32b}.

Reinforcement learning (RL) has been shown to enable this behavior as training progresses, with key "aha" moments in the training dynamics where models begin to generate longer responses and spontaneously develop alternative strategies for problem-solving, verification, and self-correction. As a result, average response length tends to grow proportionally with performance~\citep{deepseekai2025deepseekr1incentivizingreasoningcapability,zeng2025simplerl,openr1}.

At the same time, contradictory signals have emerged around whether RL is strictly necessary to enable these behaviors.
Cost-effective approaches suggest that access to long reasoning traces may be the key. In fact, recent work shows it is possible to replicate or sometimes even surpass o1 and R1 performance on challenging math benchmarks using long reasoning traces and supervised fine-tuning 
~\citep{muennighoff2025s1simpletesttimescaling, openthoughts}.

This growing belief—that longer reasoning traces equals better reasoning—has shaped much of the recent progress in training and scaling strategies. However, is longer thinking always better? At the surface level, it may appear so.
Long thought allows the model to explore alternative solutions paths, define subgoals, backtrack, verify and self-correct. These cognitive behaviors, akin to human problem-solving, have been indeed shown to be beneficial for reasoning models~\citep{gandhi2025cognitivebehaviorsenableselfimproving}. Furthermore, it is intuitive that complex problems inherently require lengthier deliberations.
However, several recent works have demonstrated that longer responses do not always yield better results. In fact, incorrect responses often involve longer reasoning traces marked by frequent switches between different lines of thought where the model prematurely abandons promising directions—a tendency coined by \cite{wang2025thoughts} as \textit{under-thinking}.
On the other hand, \textit{over-thinking} occurs when the model inefficiently expends resources by engaging in excessive verification or redundant checks after arriving at a final answer, contributing minimally to accuracy improvements \cite{chen2024not}.

Then, is shorter necessarily better?
The phenomena of under-thinking and over-thinking have motivated several ad-hoc heuristics that use response length as a proxy for downstream performance~\citep{wang2025thoughts, fu2024efficiently}.
For instance, a naive approach to boost a model’s reasoning capability is supervised fine-tuning on the shortest reasoning trajectories distilled from large state-of-the-art models such as DeepSeek-R1 671B.
 However,
blind shortening is  inherently limited, as length alone may not reliably indicate thoughtfulness or reasoning quality. Short responses may overlook nuanced considerations or miss essential parts of the meta-thinking process~\citep{Xiang2025TowardsS2}. Furthermore, employing simple length-based heuristics disregards the complexity and semantic coherence of generated content, potentially discarding useful reasoning sequences that are verbose yet insightful.

Our goal is to consolidate these disparate observations on the quality of reasoning trajectories. We ask---if overly long reasoning is not always beneficial, and blind shortening is suboptimal, how can we discourage under-thinking and over-thinking, and collect more efficient and effective solutions?
We argue that search is an effective means of eliciting better reasoning-producing trajectories that are both efficient and insightful, yet shorter in length---and can be used to train stronger student models.


In this work, we introduce \textit{\ouralgo}, a search algorithm in the spirit of Monte-Carlo Tree Search (MCTS) for distilling higher quality reasoning data from large reasoning models. \textit{\ouralgo} retrospectively revises a given reasoning path by suppressing unnecessary thought switches to collect more efficient and effective alternatives. 
Figure~\ref{fig:enter-label} shows an example of \textit{\ouralgo} refining a reasoning trace from DeepSeek-R1. It expands promising thoughts that were prematurely abandoned to mitigate under-thinking while pruning redundant thoughts once the correct answer becomes evident to reduce over-thinking, resulting in more effective yet shorter reasoning traces.



Contrary to prior attempts where search struggled to improve reasoning effectively, we show that our method is highly effective in two key settings:
\textbf{(1) Self-improvement}---\ouralgo can bootstrap self-improvement in reasoning models, by training a model on its own \textit{\ouralgo}-ed trajectories. We demonstrate that this simple step, despite not relying on frontier model capabilities,  yields significant performance gain (of up to 7.7\%) while reducing inference time by 31.2\%.
\textbf{(2) Weak-to-strong revision}---\textit{\ouralgo} can revise even the reasoning traces generated by an expensive, frontier reasoning model with a substantially smaller, more efficient model, yet significantly improving the quality of dataset. For example, we revise reasoning traces generated by R1-671B using a 20× smaller model R1-distill-32B as the \textit{\ouralgo}-er. 
Yet after training on this revised data, Qwen2.5-32B achieves performance comparable to R1-distill-32B, yielding an 11.3\% reduction in reasoning length and a 2.4\% performance improvement compared to fine-tuning on the original R1-671B's trajectories. And, more excitingly, R1-distill-7B and R1-distill-32B, fine-tuned on this revised data, achieve new state-of-the-art reasoning performance at the 7B and 32B scales while yielding the highest inference time efficiency.

\section{Method}\label{sec:method}
We introduce \textit{\ouralgo}, an MCTS-inspired algorithm that explores untaken steps for deeper and more efficient reasoning.
Its goal is to revise and improve a given reasoning path by encouraging continuation instead of prematurely switching to a new thought, ultimately seeking to reach the correct answer more efficiently, i.e. with fewer steps.

\begin{figure}[t!]
    \centering
    \includegraphics[width=0.99\textwidth]{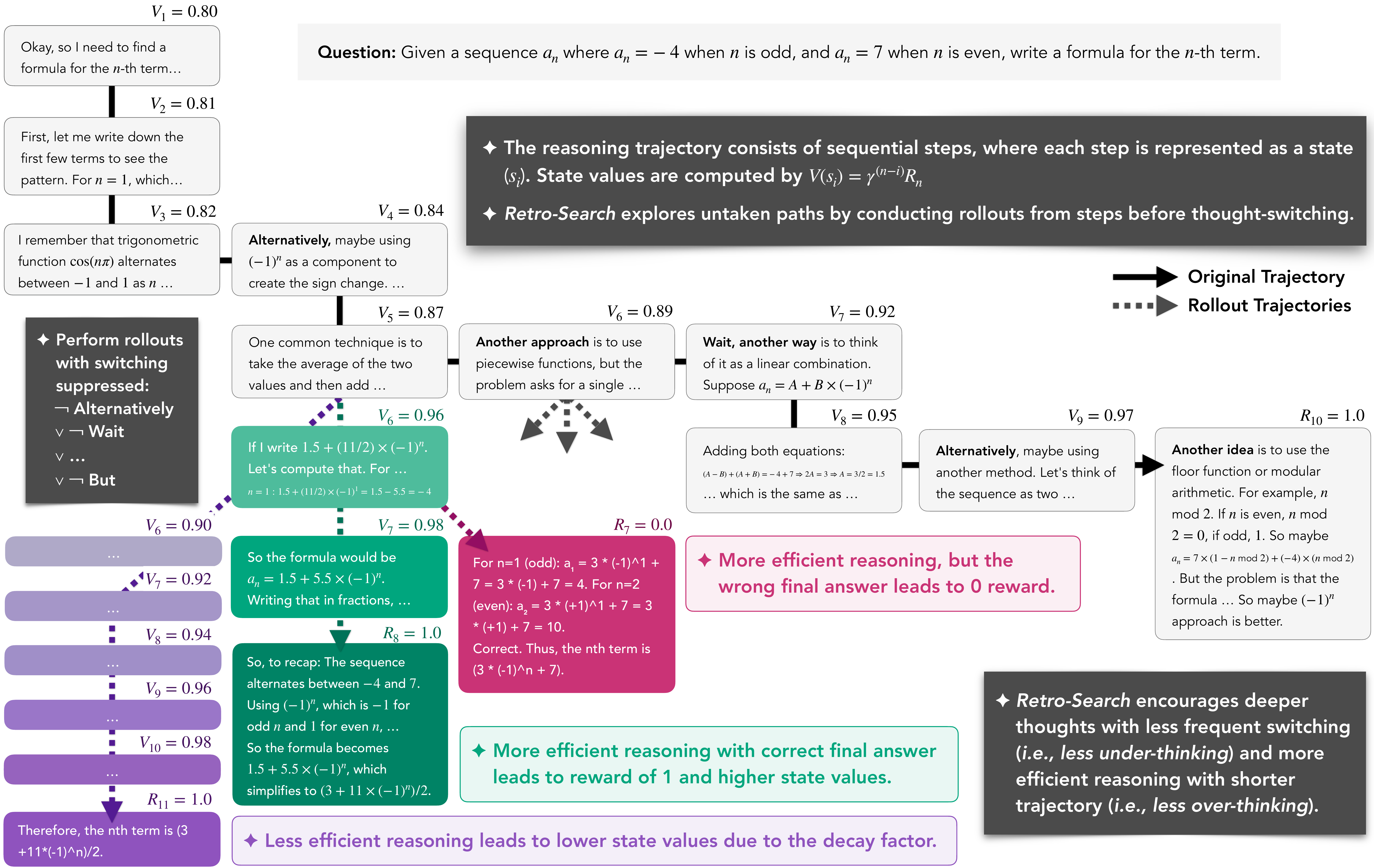}
    \vspace{-3pt}
    \caption{
        An overview of our \textit{\ouralgo} algorithm.
        The algorithm iterates through the thoughts and explores untaken paths from steps that come before a thought-switch, which is marked by transition keywords like "wait" or "another approach."
        During the process, it performs multiple rollouts, suppressing these transition keywords in the immediate next step.
        If the search is successful, the existing trajectory is replaced with the new rollout, and the process continues through the updated trajectory.
    }
    \label{fig:method}
    \vspace{-5pt}
\end{figure}

\subsection{Preliminaries}\label{sec:sub_prelim}

Consider a reasoning model $\model$ that, when given a question $\question$, generates both an intermediate reasoning \textbf{trajectory} $\reasoning$ and a final solution $\answer$.
Formally, given an input question $\question \in \questionspace$, the model $\model$ produces $(\reasoning, \answer) := \model(\question)$,
where $\reasoning \in \reasoningspace$ denotes the chain of reasoning, or chain of “thoughts”, and $\answer \in \answerspace$ represents the final solution to $\question$.

Each trajectory $\reasoning$ can be decomposed in to a set of \textbf{thoughts}, \textit{i.e.,} $\reasoning:= \{\thought 1, \thought 2, \ldots, \thought \tau\}$. Each $\thought \tau$ denotes an individual thought, and each thought may perform distinctive role such as trying out a new solution strategy, reflecting its progress, back-tracking or verifying calculations, etc. In order to differentiate between independent thoughts, we attend to the fact that models often leverage \textbf{transition keywords} (e.g., ``alternatively'') to make a natural transition between thoughts, e.g. $\thought \tau \rightarrow \thought {\tau+1}$. We utilize these linguistic markers to segment and extract individual thoughts from the full reasoning trace.

Each thought $\thought \tau$ itself is a sequence of \textbf{intermediate steps} $\stepinthought \tau i$s---that is, $\thought \tau:= \{ \stepinthought \tau 1, \stepinthought \tau 2, ...., \stepinthought \tau k \} $. These intermediate steps $\stepinthought{\tau}{k}$ represent atomic units of reasoning within a thought—such as sub-conclusions, calculations, or logical deductions.
In practice, steps are delimited by '\texttt{\textbackslash n\textbackslash n}' (double newline) characters in the model's output.
We adapt the convention of using the superscript $\tau$ to index the thought, and the subscript $k$ to index the step within that thought. For example, $\stepinthought{\tau}{k}$ refers to the $k$-th step within the $\tau$-th thought.

Utilizing the notations above, we represent a complete reasoning trajectory $T$ as:
\begin{equation}\label{eq:complete_reasoning_trajectory}
T= \bigg\{\{ \stepinthought 1 1, \stepinthought 1 2, \ldots, \stepinthought 1 {k_1} \}, \{ \stepinthought 2 1, \stepinthought 2 2, \ldots, \stepinthought 2 {k_2} \}, \ldots, a \bigg\}
\end{equation}
\textbf{The under-thinking issue: too many shallow thoughts.} \hspace{1mm} Previous studies have shown that R1-like models exhibit an under-thinking issue in their reasoning process~\citep{wang2025thoughts}. These models frequently abandon promising reasoning paths prematurely, leading to inadequate depth of reasoning on challenging problems. This phenomenon (1) occurs more frequently on harder problems, (2) leads to frequent switching between different thoughts without reaching a conclusion in each, and (3) correlates with incorrect responses due to insufficient exploration of reasoning paths.

\textbf{The over-thinking issue: too many redundant thoughts.} \hspace{1mm} Conversely, R1-like models also suffer from an over-thinking issue ~\citep{sui2025stopoverthinkingsurveyefficient,chen2024not}, where they expend excessive compute on questions that are exceptionally simple or for which the answer is already evident. The model tends to generate unnecessary thoughts such as self-doubt and redundant verification, even when it produces the correct answer within its early steps.

The seemingly contradictory issues of under-thinking and over-thinking share a common cause: unnecessarily initiating a new thought.
In under-thinking, the model switches to a new thought without fully exploring a previously promising path.
In over-thinking, despite the answer being evident, a new thought is started instead of directly generating the answer.

\subsection{\ouralgo}\label{sec:sub_retro_search}


The goal of \textit{\ouralgo} is to start from a tuple $(\question, \reasoning)$ generated by $\mathcal{M}$, and search for an improved trajectory $\betterreasoning$ using a revision model $\widehat{\mathcal{M}}$. 
Here, we focus only on revising $\reasoning$ that leads to the correct final answer (i.e., $\answer = \answer^\star$).
Intuitively, we consider $\betterreasoning$ to be better than $\reasoning$ if it leads to the same final answer $\answer$ with fewer reasoning steps---i.e., by avoiding both over-thinking and under-thinking. We specifically consider two settings of \textit{\ouralgo}, depending on how we set the revision model---(1) \textbf{Self-\ouralgo}, where $\widehat{\mathcal{M}}$ is set to be the original model $\mathcal{M}$ that produced $\reasoning$, and (2) \textbf{Weak-to-Strong-\ouralgo (W2S-\ouralgo)}, where $\widehat{\mathcal{M}}$ is a smaller, cost-efficient model than $\mathcal{M}$.

\paragraph{Collecting alternative rollouts}
The core rationale behind \textit{\ouralgo} is that there may exist an alternative trajectory for a given problem that is shorter than the original trajectory, yet still leads to a correct answer.
To discover such a trajectory, we iteratively explore alternative rollouts to investigate what would happen if, instead of starting a new thought $\thought {\tau +1}$ after $\thought \tau$ (i.e., generate $s^{\tau+1}_1$), we continued the current thought $\thought \tau$.
Concretely, for each thought $\thought \tau$ in $T$ (Eq. \ref{eq:complete_reasoning_trajectory}), we generate an alternative rollout using $\widehat{\mathcal{M}}$ as:
\begin{align}
\{ \stepinthought \tau {k+1},\ldots, \answer \} \sim \widehat{\model}\left( \thought 1, \thought 2, ..., \{ \stepinthought \tau 1, \stepinthought \tau 2, \ldots, \stepinthought \tau {k} \}\right)
\end{align}
Importantly, when generating the immediate next step $\stepinthought \tau {k+1}$, we constrain the model to stay within a single thought by preventing it from initiating a new one in the next step---by prohibiting the usage of thought-transition keywords (e.g., ``alternatively,'' ``wait'') during the decoding process.
This encourages deeper exploration of the current thought rather than prematurely switching to different lines of thought.
Subsequent steps after $\stepinthought \tau {k+1}$ are generated without constraints to allow free on-policy exploration.

\paragraph{Evaluating alternative rollouts} To determine whether the alternative rollout $\{ \stepinthought \tau {k+1},\ldots, \answer \}$ is better than the existing path $\{ \stepinthought {\tau+1} {1},\ldots, \answer \}$, we define a value function $V(s)$ over the $i$-th step $s_i$ in the trajectory $\{ s_1,\ldots, \answer \}$ to compare $V(\stepinthought \tau {k+1})$ with $V(\stepinthought {\tau+1} {1})$:
\begin{align} 
V(s_i, \answer^\star) := \gamma^{N-i}  R(\answer(s_i), \answer^\star) 
\end{align}
where $N$ represents the total number of steps in the trajectory $\{ s_1,\ldots, \answer \}$. Here, we write $\answer(s_i) := \{s_i, \ldots, \answer\}$ to explicitly emphasize that the value depends on the specific step $s_i$ and captures the autoregressive dependence of the generated answer $\answer$ on the continuation from step $s_i$.
The reward function $R(a, a^\star)$ is binary, indicating whether the generated answer $a$ matches the ground truth $a^\star$ (i.e., using a verifiable reward).
We apply a decay factor $\gamma$ to discount future rewards, assigning higher value to paths that reach the correct answer in fewer steps.
Concretely, we set to $\gamma = 0.9$ in our experiments.
In what follows, we drop the detailed notation and refer to the above simply as $V(s)$ for clarity.

If $V(\stepinthought \tau {k+1}) > V(\stepinthought {\tau + 1} 1)$, the rollout reaches the correct final answer in fewer steps, and we replace the existing path $\{ \stepinthought {\tau+1} {1},\ldots, \answer \}$ with the rollout $\{ \stepinthought \tau {k+1},\ldots, \answer \}$.
This could occur when exploring deeper along the current thought is more effective, thus reducing under-thinking.
 Alternatively, $\stepinthought \tau {k+1} = \answer$ indicates that the previous thought steps are already sufficient for the model to generate the correct solution directly, thereby reducing over-thinking.


In contrast, if $V(\stepinthought \tau {k+1}) < V(\stepinthought {\tau + 1} 1)$, the existing path is better. The alternative path either reaches a wrong answer or reaches the correct answer with more steps. This suggests that switching to a new thought was effective and necessary, and thus the existing transition should be preserved.
In practice, we sample multiple alternative rollouts (two in our experiments) and retain the best one—that is, the rollout with the highest value.
We then proceed to examine the next thought in the updated reasoning trajectory.
Please see Figure~\ref{fig:method} for a concrete example, and Algorithm~\ref{alg:alg} in Appendix~\ref{pseudo-code} for more details.

\vspace{-2mm}
\paragraph{Retro-Search with Partial Revisions}
We also propose a more computationally efficient variant of \textit{\ouralgo}.
Instead of iteratively applying the revision procedure starting from the first thought, this version randomly samples a position in the trajectory at which to begin the revision.
This is particularly useful when revising with larger models—for instance, the R1-32B model in our setting—where full iterative revision would be prohibitively expensive.
\vspace{-1mm}
\section{Experiments}\label{sec:exp}



\vspace{-1mm}
\subsection{Setup}
\vspace{-1mm}
\paragraph{Data Generation} 
We use 40K math questions from NuminaMath~\citep{numina_math_datasets}.
Specifically, we sample NuminaMath questions from \texttt{OpenThoughts-114k}\footnote{\href{https://huggingface.co/datasets/open-thoughts/OpenThoughts-114k}{https://huggingface.co/datasets/open-thoughts/OpenThoughts-114k}} dataset, which is the training data of OpenThinker-7B and OpenThinker-32B models. We experiment with two settings: 1) \textbf{Self-Retro-R1-7B}, where we first generate responses using the R1-distilled Qwen2.5-7B model and then revise them with the same model as the \textit{\ouralgo}-er. 2) \textbf{W2S-Retro-R1-32B}, where we take responses from the DeepSeek-R1 671B model in the \texttt{OpenThoughts} dataset and revise them using a weaker model, R1-distilled Qwen2.5-32B.  More details are in Appendix~\ref{app:data}.

\vspace{-2mm}
\paragraph{Model Training} 
We trained four models using data generated by \textit{\ouralgo}: Qwen2.5-7B-Instruct, R1-distilled Qwen2.5-7B, Qwen2.5-32B-Instruct and R1-distilled Qwen2.5-32B with supervised fine-tuning.
All models are fine-tuned for five epochs with learning rate of 1e-5, and sequence length of 16K. More details are in Appendix~\ref{app:training}.

\vspace{-2mm}
\paragraph{Baselines} 
We compare our trained models with a total of eleven open-weight models across two model size categories --- six 7B models and five 32B models. These include instruction-tuned models such as Qwen2.5-7B-Inst \citep{yang2024qwen2}, Qwen2.5-Math-7B, Qwen2.5-Math-7B-Inst \citep{yang2024qwen2math} and Qwen2.5-32B-Inst \citep{yang2024qwen2}, as well as reasoning models such as OpenR1-Qwen-7B \citep{openr1}, OpenThinker-7B \citep{openthoughts},  R1-distill Qwen2.5-7B \citep{deepseekai2025deepseekr1incentivizingreasoningcapability}, OpenThinker-32B \citep{openthoughts}, QwQ-32B-Preview \citep{qwq32b}, Sky-T1-32B-Preview \citep{sky-t1}, and R1-distill Qwen2.5-32B~\citep{deepseekai2025deepseekr1incentivizingreasoningcapability}.
More baseline details are in Appendix~\ref{app:baselines}.

\vspace{-2mm}
\paragraph{Benchmarks and Metrics} 
We evaluate models on seven math-specific benchmarks: AIME25, AIME24, AMC23, GaoKao23English~\citep{zhong2023agieval}, OlympiadBench~\citep{he2024olympiadbench}, GSM8K~\citep{cobbe2021training}, and MATH500~\citep{letsverifystepstep}.
The first five benchmarks focus on olympiad-level math problems, where AIME25 and AIME24 each contain 30 problems and AMC23 contains 40 problems. GSM8K includes grade school math problems, and MATH500 includes high-school math competition problems.

For evaluation, we report two metrics: \textit{accuracy} to measure the performance, and average \textit{response length} to measure computational efficiency during inference.
For accuracy, we use exact match between the model's prediction and the reference answer, with Qwen's official implementation\footnote{\href{https://github.com/QwenLM/Qwen2.5-Math/tree/main}{https://github.com/QwenLM/Qwen2.5-Math/tree/main}.} for answer verification.\footnote{Note that evaluation results can significantly vary depending on the specifics of the answer verification, so we recommend to use the same implementation for reproduction.}
For response length, we tokenize the responses using the Qwen2.5-7B-Instruct tokenizer and compute the number of output tokens.

Metrics are computed individually for each benchmark and then averaged using macro averaging
to produce the final scores. Since there is no universally optimal decoding strategy that works well across all models, we report results under two commonly used decoding setups: greedy decoding (T=0), following \citet{muennighoff2025s1simpletesttimescaling}, and temperature sampling (T=0.6 with top-p=0.95), following \citet{deepseekai2025deepseekr1incentivizingreasoningcapability}.
We took an average of results from five different seeds for the temperature sampling setup. In Appendix~\ref{app:evals}, we share the full results including the confidence interval of the results.

\subsection{Evaluation Results}
\paragraph{Self \ouralgo teaches stronger and more efficient student models than vanilla data generation.}

\begin{table}[t!]
\centering
\small
\resizebox{0.85\textwidth}{!}{
\begin{tabular}{lllll}
\toprule
& \multicolumn{2}{c}{Greedy Decoding} & \multicolumn{2}{c}{Sampling (T=0.6, p=0.95)} \\
\cmidrule(r{0.25em}l{0.25em}){2-3} \cmidrule(r{0.25em}l{0.25em}){4-5}
Models & \multicolumn{1}{c}{Accuracy ($\uparrow$)} & \multicolumn{1}{c}{Length ($\downarrow$)} & \multicolumn{1}{c}{Accuracy ($\uparrow$)} & \multicolumn{1}{c}{Length ($\downarrow$)} \\
\midrule
Baselines (7B) & & & & \\
\cmidrule(l{0.5em}r{0.5em}){1-1}
Qwen2.5-Math-7B & 41.1 & 1182 & 39.0 & 1225 \\
Qwen2.5-Math-7B-Inst & 53.1 & 982 & 52.7 & 985 \\
OpenR1-Qwen-7B & 67.6 & 9463 & 71.7 & 7740 \\
OpenThinker-7B & 53.8 & 14477 & 59.1 & 9835 \\
\midrule
Qwen2.5-7B-Inst & 48.7 & 985 & 47.9 & 1033 \\
\cmidrule(l{0.6em}r{0.7em}){1-1}
~+ R1-7B & 49.7 & 14365 & 55.4 & 8959 \\
\textbf{~+ Self-Retro-R1-7B} & 51.7 (\textcolor{blue}{$+$4.1\%}) & 11050 (\textcolor{blue}{$-$23.1\%}) & 55.8 (\textcolor{blue}{$+$0.7\%}) & 8263 (\textcolor{blue}{$-$7.8\%})\\
\cmidrule(l{0.6em}r{0.7em}){1-1}
~+ R1-671B & 51.5 & 14302 & 58.4 & 9824 \\
\textbf{~+ W2S-Retro-R1-32B} & 55.3 (\textcolor{blue}{$+$7.3\%}) & 13569 (\textcolor{blue}{$-$5.1\%}) & 57.8 (\textcolor{gray}{$-$1.1\%}) & 8940 (\textcolor{blue}{$-$9.0\%}) \\
\midrule
R1-distill-Qwen2.5-7B & 64.5 & 10600 & 71.0 & 6831 \\
\cmidrule(l{0.6em}r{0.7em}){1-1}
~+ R1-671B & 68.4 & 9418 & 71.7 & 7172 \\
~+ \textbf{W2S-Retro-R1-32B} & 70.8 (\textcolor{blue}{$+$3.5\%})  & 8800 (\textcolor{blue}{$-$6.6\%}) & \textbf{73.1} (\textcolor{blue}{$+$2.0\%}) & \textbf{6535} (\textcolor{blue}{$-$8.9\%}) \\
 \midrule
 \midrule
Baselines (32B) & & & & \\
\cmidrule(l{0.5em}r{0.5em}){1-1}
OpenThinker-32B & 73.0 & 8001 & 75.9 & 6840 \\
QwQ-32B-Preview & 70.9 & 5164 & 68.3 & 5163 \\
Sky-T1-32B-Preview & 62.0 & 2367 & 62.9 & 2018 \\
\midrule
Qwen2.5-32B-Inst & 56.1 & 975 & 55.9 & 761 \\
\cmidrule(l{0.6em}r{0.7em}){1-1}
~+ R1-671B & 76.2 & 7074 & 75.6 & 6676 \\
\textbf{~+ W2S-Retro-R1-32B} & 74.6 (\textcolor{gray}{$-$2.2\%}) & 6809 (\textcolor{blue}{$-$3.7\%}) & 77.5 (\textcolor{blue}{$+$2.4\%}) & 5923 (\textcolor{blue}{$-$11.3\%})\\
\midrule
R1-distill Qwen2.5-32B & 73.1 & 8566 & 77.7 & 6173 \\
\cmidrule(l{0.6em}r{0.7em}){1-1}
~+ R1-671B (12K) & 80.4 & 6470 & 79.8 & 6164 \\
\textbf{~+ W2S-Retro-R1-32B (12K)} & 79.9 (\textcolor{gray}{$-$0.6\%}) & 6091 (\textcolor{blue}{$-$5.9\%}) & \textbf{81.0} (\textcolor{blue}{$+$1.5\%}) & \textbf{5301} (\textcolor{blue}{$-$14.0\%})\\
\bottomrule
\end{tabular}}
\caption{\textbf{\ouralgo provides better training data.} Model evaluation results averaged across seven math benchmarks (AIME25, AIME24, AMC23, GaoKao23English, OlympiadBench, GSM8K, and MATH500). We report results from two setups: greedy decoding (T = 0) and temperature sampling (T = 0.6 with top-p = 0.95). $+ X$ indicates that the model is fine-tuned with data X. Only when fine-tuning R1-distill Qwen2.5-32B, we used 12K instances, as using more data did not improve results. The results indicate that: (1) models trained with \textit{\ouralgo} data are more computationally efficient during inference while generally showing better performance; and (2) 
weak-to-strong \textit{\ouralgo} enables new SOTA at 7B and 32B scales.}

\label{tab:results}
\end{table}
\begin{table}[t!]
\centering
\small
\setlength{\tabcolsep}{12pt}
\resizebox{0.85\textwidth}{!}{
\begin{tabular}{lcccc}
\toprule
 & \multicolumn{2}{c}{Greedy Decoding} & \multicolumn{2}{c}{Sampling (T=0.6, p=0.95)} \\
\cmidrule(r{0.25em}l{0.25em}){2-3} \cmidrule(r{0.25em}l{0.25em}){4-5}
Qwen2.5-7B-Inst & \multicolumn{1}{c}{Accuracy ($\uparrow$)} & \multicolumn{1}{c}{Length ($\downarrow$)} & \multicolumn{1}{c}{Accuracy ($\uparrow$)} & \multicolumn{1}{c}{Length ($\downarrow$)} \\
\midrule
~+ R1-7B & 49.7 & 14365 & 55.4 & 8959 \\
~+ R1-7B-Shortest & 50.3 & 12340 &  54.6 & \textbf{8009}\\
\textbf{~+ Self-Retro-R1-7B} & \textbf{51.7} & \textbf{11050} & \textbf{55.8} & 8263 \\
\bottomrule
\end{tabular}
}
\caption{\textbf{Simply selecting the shortest path for training is suboptimal for model accuracy.} We fine-tuned Qwen2.5-7B-Inst with different training data and compare results. We sample eight responses using R1-distilled Qwen2.5-7B and choose the shortest response.}
\label{tab:shortest}
\end{table}
We compare fine-tuning the student model, Qwen2.5-7B-Instruct, using data from our \textit{Self-Retro-R1-7B} against fine-tuning with data sampled from the R1-distilled Qwen2.5-7B model before revision, referred to as \textit{R1-7B} in Table~\ref{tab:results}. Compared to models trained on \textit{R1-7B}, the model trained on \textit{Self-Retro-R1-7B} produces responses that are 23.1\% shorter while improving accuracy by +4.1\% under greedy decoding.

We further compare \textit{\ouralgo} against another baseline, \textit{R1-7B-Shortest}, which selects the shortest response for model training after sampling eight responses per questions using R1-distilled Qwen2.5-7B. As shown in Table~\ref{tab:shortest}, although training with the shortest response can enhance efficiency when compared to R1-7B, it does not improve the model performance as much as our \ouralgo, clearly demonstrating the effectiveness of our \ouralgo.

\vspace{-2mm}
\paragraph{Weak-to-Strong \ouralgo enables new SOTA reasoning models at 7B and 32B scales, excelling in both performance and efficiency.}
While \textit{Self-Retro} has proven effective, using a large model such as DeepSeek-R1-671B for both generation and revision is computationally implausible. 
We evaluate the effectiveness of weak-to-strong revision, where DeepSeek-R1-671B's generations are \textit{\ouralgo}-ed by R1-distilled Qwen2.5-32B, denoted as \textit{W2S-Retro-R1-32B}. We fine-tune student models on this data and compare them to those fine-tuned on unrevised data from DeepSeek-R1-671B, referred to as \textit{R1-671B} in Table~\ref{tab:results}.


\textit{W2S-Retro-R1-32B} proves to be effective, enabling new SOTA reasoning models at 7B and 32B scales. We fine-tuned four models --- Qwen2.5-7B-Instruct, R1-distilled Qwen2.5-7B, Qwen2.5-32B-Instruct and R1-distilled Qwen2.5-32B ---- and consistently observed reduced response lengths and improved performance across different setups compared to models fine-tuned on \textit{R1-671B}.
Surprisingly, R1-distilled Qwen2.5-7B and R1-distilled Qwen2.5-32B fine-tuned on \textit{W2S-Retro-R1-32B}, achieve new SOTA reasoning performance in the sampling setting at the 7B and 32B scales, while yielding the highest inference time efficiency.
In addition, Qwen2.5-32B fine-tuned on \textit{W2S-Retro-R1-32B}, achieves performance comparable to R1-distill-32B, yielding an 11.3\% reduction in reasoning length and a 2.4\% performance improvement compared to fine-tuning on the \textit{R1-671B} data. 
Notably, it also outperforms OpenThinker-32B in accuracy 
while being more efficient (13.4\%--14.9\% shorter response). This is particularly significant given that OpenThinker-32B is trained on around 2.5 times more data than our \textit{W2S-Retro-R1-32B} and use DeepSeek-R1 671B 
for response generation.



\vspace{-2mm}
\paragraph{\ouralgo enables self-improvement of R1-distilled models.}

\begin{table}[t!]
\centering
\small
\setlength{\tabcolsep}{10pt}
\resizebox{0.85\textwidth}{!}{
\begin{tabular}{lllll}
\toprule
& \multicolumn{2}{c}{Greedy Decoding} & \multicolumn{2}{c}{Sampling (T=0.6, p=0.95)} \\
\cmidrule(r{0.25em}l{0.25em}){2-3} \cmidrule(r{0.25em}l{0.25em}){4-5}
Models & \multicolumn{1}{c}{Accuracy ($\uparrow$)} & \multicolumn{1}{c}{Length ($\downarrow$)} & \multicolumn{1}{c}{Accuracy ($\uparrow$)} & \multicolumn{1}{c}{Length ($\downarrow$)} \\
\midrule
R1-distill Qwen2.5-7B & 64.5 & 10600 & 71.0 & 6831 \\
\textbf{~+ Self-Retro-R1-7B} & 69.5 (\textcolor{blue}{$+$7.7\%}) & 7295 (\textcolor{blue}{$-$31.2\%}) & 70.6 (\textcolor{gray}{$-$0.6\%}) & 5406 (\textcolor{blue}{$-$20.9\%})\\
\bottomrule
\end{tabular}
}
\caption{\textbf{\ouralgo allows self-improvement of the models.} Fine-tuning the R1-distilled Qwen2.5-7B model with self-revision data (\textit{Self-Retro-R1-7B}) significantly improves efficiency, while maintaining or even improving accuracy.}
\label{tab:selfimp}
\end{table}

We fine-tune the R1-distilled Qwen2.5-7B model with our \textit{Self-Retro-R1-7B}.
Results in Table~\ref{tab:selfimp} show significant accuracy improvement ($+$7.7\%) and response length reduction (31.2\%) for greedy decoding, compared to R1-distill Qwen2.5-7B. 
There is a small performance reduction for temperature sampling ($-$0.6\%), but the length reduction is substantial (20.9\%).
As Self-Retro-R1-7B uses R1-distilled Qwen2.5-7B model for response generation, revision, and fine-tuning the model itself, this shows the self-improvement capabilities enabled by \ouralgo.

\subsection{Analyses}\label{sec:analyses}

\begin{table}[h]
    \centering
    \small
    \begin{adjustbox}{width=\linewidth}
    \begin{tabular}{lcccccc}
        \toprule
        & \multicolumn{3}{c}{Synthesized Training Data} & \multicolumn{3}{c}{Student Model’s Reasoning Trace} \\
        \cmidrule(lr){2-4} \cmidrule(lr){5-7}
                            & \makecell{\#Transition \phantom{ ($\downarrow$)}\\Keywords ($\downarrow$)}   & \makecell{\#Steps/Thought \\  ($\uparrow$) } & \makecell{Relative Location\\of Solution  ($\uparrow$)}   & \makecell{\#Transition \phantom{ ($\downarrow$)}\\Keywords ($\downarrow$)}   & \makecell{\#Steps/Thought \\  ($\uparrow$) }   & \makecell{Relative Location\\of Solution ($\uparrow$)}\\
        \midrule
        R1-7B               & 85.9          & 3.7               & 0.67                      & 229.2          & 4.7           & 0.59  \\
        \textbf{Self-Retro-R1-7B}    & \textbf{32.7}          & \textbf{5.3}             & \textbf{0.73}                      & \textbf{183.2}           & \textbf{5.4}          & \textbf{0.64} \\
        \midrule                                                             
        R1-671B             & 35.3          & 3.8               & 0.59                      & 80.0          & 3.0            & 0.44  \\
        \textbf{W2S-Retro-R1-32B}    & \textbf{10.4}           &\textbf{4.9}               & \textbf{0.60}                     & \textbf{70.1}          & \textbf{3.2}            & \textbf{0.48}  \\
        \bottomrule
    \end{tabular}
    \end{adjustbox}
    \caption{
        The average number of transition keywords, the number of steps per thought, and the relative location of the first appearance of the solution in the reasoning trajectory are taken from both the training data and the fine-tuned student model, Qwen2.5-7B.
    }
    \label{tab:analyses}
\end{table}

We quantitatively analyze the reasoning trajectories in the synthesized training data using our \textit{\ouralgo}, as well as those generated by the fine-tuned student model Qwen2.5-7B.
Table \ref{tab:analyses} reports the average number of transition keywords, number of steps per thought, and the relative location where the solution first appears in the trajectory (with values closer to 1 indicating that the solution is nearer the end).
The synthesized reasoning traces from \textit{\ouralgo} contain significantly fewer transition keywords than those from R1-7B and R1-671B.
As a result, thoughts from \textit{\ouralgo} include more steps than those from R1-7B and 671B, indicating deeper thoughts.
Additionally, the solution tends to appear later in the trajectory, suggesting that our approach show less redundant thoughts after the final solution is derived.
These trends are also consistent in the reasoning outputs from the student model, showing that \textit{\ouralgo} reduces both under-thinking and over-thinking.

\vspace{-1mm}
\section{Related Works}\label{sec:related_works}
\vspace{-1mm}
Test-time compute has emerged as a new axis of scaling for LLM reasoning. While prior research in this direction have focused on parallel scaling---repeated sampling of trajectories followed by aggregation \citep{large_language_monkeys, snell2024scalingllmtesttimecompute, rebase}, recent efforts have focused on sequential scaling---where models are trained to back-track, evaluate, and revise its thought by generating a long, monolithic CoT. Representative models such as O1 and R1 \citep{o1, deepseekai2025deepseekr1incentivizingreasoningcapability} are trained via large-scale reinforcement learning, demonstrating that models can learn to generate long CoTs without relying on bespoke reward models \citep{letsverifystepstep, lessonsdevelopingprocessreward}, or tree search \citep{feng2024alphazeroliketreesearchguidelarge, rest-mcts}. Subsequent projects in open-source community aim to replicate these reasoning models \citep{openr1, o1_replication_journey}. These works often utilize frontier reasoning models to generate synthetic long thought traces, and showing suprising gain in reasoning capabilities via simple supervised fine-tuning \citep{openr1, sky-t1, muennighoff2025s1simpletesttimescaling}. Our work builds upon these prior efforts, focusing on (1) better-quality reasoning paths by targeted revision of verbose sub-traces, and (2) demonstrating self-improvement beyond typical strong-to-weak distillation, where smaller models can self-improve in both performance and efficiency.


Meanwhile, concurrent works reveal limitations of reasoning models in their in-efficiency of test-time scaling. Longer generation does not necessarily correlate with better accuracy \citep{zeng2025revisitingtesttimescalingo1like}, and in practice, shorter trajectories are more likely to be correct. Models tend to overthink \citep{danger-of-overthinking, sui2025stopoverthinkingsurveyefficient, chen2024not}, \textit{i.e.,} they generate unnecessarily long trajectgories that do not contribute to the performance. Models also exhibit underthinking \citep{wang2025thoughts}---while they appear to explore diverse plausible paths, models often switch between paths without sufficient exploration on one path. \citet{more-is-less} suggests the source of inefficiency may lie in the regularities of the training data we use, and theoretically show that training on CoTs that are longer than the optimal length for the model can hurt its performance. Several measures have been proposed to mitigate these findings, such as auxiliary learnable parameters  \citep{bao2025learningstopoverthinkingtest, zhang2025lightthinkerthinkingstepbystepcompression}, calibration \citep{self-calibration}, and decoding-time algorithm \citep{chain_of_draft, wider_or_deeper}.  \ouralgo aligns with these prior efforts, and importantly revisits the value of search algorithm in improving both the efficiency and performance of test-time scaling.

\vspace{-1mm}
\section{Conclusions}\label{sec:conc}
\vspace{-1mm}
In this work, we introduced \textit{\ouralgo}, a novel algorithm for synthesizing reasoning data designed to equip reasoning models with efficient (shorter average response length) and effective (higher accuracy) test-time scaling.
Inspired by the MCTS algorithm, \textit{\ouralgo} retrospectively revises reasoning trajectories—eliminating unnecessary thought switches (under-thinking) and trimming redundant steps after the correct answer becomes evident (over-thinking).
Quantitatively, we show that \ouralgo is highly effective for self-improvement and weak-to-strong revision.
Specifically, R1-distill-7B, fine-tuned on its own \textit{\ouralgo}-ed traces, reduces the average reasoning length by 31.2\% while improving performance by 7.7\% across seven math benchmarks.
Notably, R1-distill-7B and R1-distill-32B, fine-tuned on weak-to-strong \textit{\ouralgo}-ed reasoning traces from R1-671B, set new state-of-the-art performance at the 7B and 32B scales while yielding the highest reasoning efficiency.
%
%
%
We hope our work reinvigorates interest in the power of search-based methods for synthetic data in reasoning models—a direction that has recently fallen out of favor, yet holds significant untapped potential.

\clearpage
\bibliography{reference}
\bibliographystyle{colm2025_conference}
\clearpage

 \begin{appendices}

\startcontents[sections]
\printcontents[sections]{l}{1}{\setcounter{tocdepth}{2}}

\newpage

\section{\ouralgo Algorithm}
\label{pseudo-code}


\begin{algorithm}[H]
\caption{\ouralgo} \label{alg:alg}
\begin{algorithmic}[1]
\small
\Require Question $q$, initial reasoning trajectory $T= \big\{\{ \stepinthought 1 1, \stepinthought 1 2, \ldots, \stepinthought 1 {k_1} \}, \{ \stepinthought 2 1, \stepinthought 2 2, \ldots, \stepinthought 2 {k_2} \}$, \ldots, a \big\}, revision model $\widehat{\mathcal{M}}$,
discount factor $\gamma$, ground truth answer $a^\star$, and reward function $R(\cdot, \cdot)$.
\Ensure Revised trajectory $\tilde{T}$ that yields answer $a^*$ with fewer steps.
\State Initialize $\tilde{T} \gets T$
\State Initialize $\thought \tau \gets \thought 1$ from $\tilde{T}$
\While{$\thought \tau$ is not the last thought in $\tilde{T}$}
    \State $\{\stepinthought \tau {k+1},\ldots, \answer \} \sim \widehat{\model}\left( \thought 1, ...,\{ \stepinthought \tau 1, \stepinthought \tau 2, \ldots, \stepinthought \tau {k} \}\right)$ \Comment{Rollout: transition keywords prohibited in $\stepinthought \tau {k+1}$}
    \State $V(\stepinthought \tau {k+1}, \answer^\star) \gets \gamma^{N-i}  R(\answer(\stepinthought \tau {k+1}), \answer^\star)$ \Comment{Compute value of the new step $\stepinthought \tau {k+1}$ (i.e., $i$-th step)}
    \If{$V(\stepinthought \tau {k+1}) > V(\stepinthought {\tau + 1} 1)$} \Comment{If the value of the new step is higher than the existing one}
        \State $\tilde{T} \gets \Big\{\thought 1, \thought 2, ...,\{ \stepinthought \tau 1, \stepinthought \tau 2, \ldots, \stepinthought \tau {k} \} \{\stepinthought \tau {k+1},\ldots, \answer \} \Big\} $ \Comment{Update the trajectory with the new rollout}
    \EndIf
    \State $\thought \tau \gets$ the next thought in $\tilde{T}$ 
\EndWhile
\State Return $\tilde{T}$
\end{algorithmic}
\end{algorithm}

\section{Data Generation Details}
\label{app:data}
When constructing \textit{Self-Retro-R1-7B}, we use the default version of \textit{\ouralgo}, whereas for \textit{W2S-Retro-R1-32B}, we use \textit{\ouralgo} with partial revision. 
When constructing \textit{Self-Retro-R1-7B}, we generate responses from R1-distill Qwen2.5-7B and filter for those with correct solutions as the base data for \textit{\ouralgo} to revise.
For \textit{W2S-Retro-R1-32B}, we directly use OpenThought data as the base, since it contains only correct responses from the DeepSeek-R1 671B model.

The transition keywords we use to segment thoughts within a reasoning trace are: 'But', 'Wait', 'Alternatively', 'However', 'Hmm', 'Hmmm', 'Not sure', 'Going back', 'Backtrack', 'Trace back', and 'Another'.

For data generation during \textit{\ouralgo}, we use top-p sampling with $p = 0.98$ and temperature $T = 1.0$ . We also tried using temperature $T = 0.6$ and found that data generated with a higher temperature tends to produce a better student model, likely due to the increased diversity in the training data induced by higher-temperature sampling. 
We set the maximum generation length to be 16384.

\section{Training Details}
\label{app:training}

We perform supervised fine-tuning of models using HuggingFace TRL~\citep{vonwerra2022trl}. For all fine-tuning experiments, we used batch size of 128, five training epochs, and cosine learning rate scheduler with warmup rate of 0.05. We used Adam optimizer with weight decay of 1e-4, with beta1=0.9 and beta2=0.95. We did not conduct hyperparameter search, so there is a potential of finding better hyperparameters. With 32 H100 GPUs, fine-tuning 7B model with 40K data took around 90 minutes, and fine-tuning 32B model took 10 hours to finish.

\section{Baselines Details}
\label{app:baselines}
For 7B models, we evaluate six open-weight models as baselines: instruction-tuned models including Qwen2.5-7B-Inst \citep{yang2024qwen2}, Qwen2.5-Math-7B, and Qwen2.5-Math-7B-Inst \citep{yang2024qwen2math}, as well as reasoning models including OpenR1-Qwen-7B \citep{openr1}, OpenThinker-7B \citep{openthoughts}, and R1-distill Qwen2.5-7B \citep{deepseekai2025deepseekr1incentivizingreasoningcapability}. These reasoning models are fine-tuned using responses from DeepSeek-R1 671B \citep{deepseekai2025deepseekr1incentivizingreasoningcapability}.
Specifically, the OpenR1-Qwen-7B model is trained on 220K math examples, with questions sourced from NuminaMath, while OpenThinker-7B is trained on the OpenThoughts-114K dataset, which includes math, science, and coding problems.

For 32B models, we evaluate five open-weight models: instruction-tuned Qwen2.5-32B-Inst \citep{yang2024qwen2}, as well as reasoning models such as OpenThinker-32B \citep{openthoughts}, QwQ-32B-Preview \citep{qwq32b}, Sky-T1-32B-Preview \citep{sky-t1}, and R1-distill Qwen2.5-32B~\citep{deepseekai2025deepseekr1incentivizingreasoningcapability}.
Both OpenThinker-32B and R1-distill Qwen2.5-32B are fine-tuned using responses generated by DeepSeek-R1 671B, with OpenThinker-32B utilizing the OpenThoughts-114K dataset.
Sky-T1-32B-Preview is trained on a 17K dataset consisting of math and coding problems, with responses generated using QwQ-32B-Preview. 
The training details of the other models are not publicly disclosed.

\section{Per-dataset Evaluation Results}
\label{app:evals}
\begin{table}[t!]
\centering
\tiny
\begin{adjustbox}{angle=270}{
\begin{tabular}{lllllllll}
\toprule
\textbf{Models (Greedy decoding, T=0)} & AIME25 & AIME24 & AMC23 & GaoKao23En & OlympiadBench & GSM8K & MATH500 & Avg. \\
\midrule
Qwen2.5-7B-Inst & 13.30 & 10.00 & 50.00 & 39.90 & 63.90 & 87.50 & 76.00 & 48.66\\
Qwen2.5-Math-7B & 6.70 & 16.70 & 55.00 & 19.60 & 47.00 & 77.30 & 65.40 & 41.10\\
Qwen2.5-Math-7B-Inst & 6.70 & 13.30 & 65.00 & 40.10 & 66.50 & 95.70 & 84.60 & 53.13\\
OpenR1-Qwen-7B & 33.30 & 46.70 & 77.50 & 54.10 & 79.70 & 92.80 & 89.20 & 67.61\\
R1-distill Qwen2.5-7B & 23.30 & 46.70 & 82.50 & 52.60 & 71.40 & 87.60 & 87.60 & 64.53\\
OpenThinker-7B & 20.00 & 20.00 & 70.00 & 40.40 & 69.10 & 78.50 & 78.40 & 53.77\\
Qwen2.5-32B-Instruct & 10.00 & 16.70 & 65.00 & 48.10 & 73.80 & 95.80 & 83.00 & 56.06\\
QwQ-32B-Preview & 40.00 & 46.70 & 82.50 & 58.70 & 81.60 & 95.20 & 91.40 & 70.87\\
Sky-T1-32B-Preview & 23.30 & 26.70 & 67.50 & 54.80 & 77.10 & 95.80 & 88.60 & 61.97\\
R1-distill Qwen2.5-32B & 46.70 & 60.00 & 90.00 & 57.20 & 76.40 & 92.90 & 88.60 & 73.11\\
OpenThinker-32B & 43.30 & 63.30 & 85.00 & 62.40 & 80.30 & 85.60 & 91.00 & 72.99\\
Qwen2.5-7B-Inst~+ R1-7B & 23.30 & 23.30 & 40.00 & 37.20 & 60.80 & 87.90 & 75.20 & 49.67\\
Qwen2.5-7B-Inst~+ R1-7B-shortest & 23.30 & 16.70 & 52.50 & 36.70 & 61.60 & 86.40 & 74.60 & 50.26\\
Qwen2.5-7B-Inst~+ Self-Retro-R1-7B & 16.70 & 20.00 & 55.00 & 38.20 & 63.90 & 87.40 & 80.60 & 51.69\\
Qwen2.5-7B-Inst~+ R1-671B & 20.00 & 16.70 & 55.00 & 41.80 & 68.80 & 78.10 & 80.00 & 51.49\\
Qwen2.5-7B-Inst~+ W2S-Retro-R1-32B & 23.30 & 20.00 & 67.50 & 42.50 & 68.10 & 85.40 & 80.00 & 55.26\\
R1-distill Qwen2.5-7B~+ R1-7B & 26.70 & 56.70 & 87.50 & 57.50 & 71.40 & 84.80 & 89.20 & 67.69\\
R1-distill Qwen2.5-7B~+ Self-Retro-R1-7B & 43.30 & 53.30 & 82.50 & 58.40 & 74.30 & 85.10 & 89.60 & 69.50\\
R1-distill Qwen2.5-7B~+ R1-671B & 36.70 & 53.30 & 82.50 & 55.70 & 78.70 & 82.10 & 89.80 & 68.40\\
R1-distill Qwen2.5-7B~+ W2S-Retro-R1-32B & 46.70 & 56.70 & 77.50 & 57.30 & 78.70 & 88.00 & 90.80 & 70.81\\
Qwen2.5-32B-Instruct~+ R1-671B & 43.30 & 73.30 & 92.50 & 62.50 & 82.10 & 86.10 & 93.60 & 76.20\\
Qwen2.5-32B-Instruct~+ W2S-Retro-R1-32B & 50.00 & 56.70 & 87.50 & 61.90 & 81.60 & 92.40 & 91.80 & 74.56\\
R1-distill Qwen2.5-32B~+ R1-671B (12k) & 70.00 & 73.30 & 97.50 & 65.50 & 81.30 & 82.80 & 92.60 & 80.43\\
R1-distill Qwen2.5-32B~+ W2S-Retro-R1-32B (12k) & 56.70 & 63.30 & 100.00 & 66.50 & 83.90 & 94.50 & 94.40 & 79.90\\
\bottomrule
\end{tabular}
}\end{adjustbox}
\caption{Per-dataset evaluation results (accuracies) using greedy decoding.}
\label{tab:appendix_finegrained_greedy_acc}

\end{table}

\begin{table}[t!]
\centering
\tiny
\begin{adjustbox}{angle=270}{
\begin{tabular}{lllllllll}
\toprule
\textbf{Models (Greedy decoding, T=0)} & AIME25 & AIME24 & AMC23 & GaoKao23En & OlympiadBench & GSM8K & MATH500 & Avg. \\
\midrule
Qwen2.5-7B-Inst & 812 & 903 & 1672 & 1591 & 645 & 253 & 1018 & 985\\
Qwen2.5-Math-7B & 1526 & 1426 & 1030 & 1715 & 1072 & 569 & 939 & 1182\\
Qwen2.5-Math-7B-Inst & 1505 & 1486 & 1157 & 1009 & 731 & 317 & 670 & 982\\
OpenR1-Qwen-7B & 16968 & 14778 & 10225 & 11643 & 5624 & 2465 & 4536 & 9463\\
R1-distill Qwen2.5-7B & 23648 & 18470 & 8519 & 13048 & 5225 & 460 & 4829 & 10600\\
OpenThinker-7B & 25417 & 26483 & 13709 & 17311 & 8021 & 2593 & 7804 & 14477\\
Qwen2.5-32B-Instruct & 1865 & 1876 & 693 & 1091 & 581 & 228 & 490 & 975\\
QwQ-32B-Preview & 10931 & 9803 & 3636 & 5768 & 2782 & 746 & 2481 & 5164\\
Sky-T1-32B-Preview & 4845 & 2870 & 3564 & 2679 & 1222 & 284 & 1107 & 2367\\
R1-distill Qwen2.5-32B & 17596 & 15200 & 7038 & 11147 & 4255 & 443 & 4285 & 8566\\
OpenThinker-32B & 14522 & 14694 & 8018 & 9577 & 3986 & 1232 & 3978 & 8001\\
Qwen2.5-7B-Inst~+ R1-7B & 25972 & 25703 & 16339 & 17954 & 7365 & 450 & 6770 & 14365\\
Qwen2.5-7B-Inst~+ R1-7B-shortest & 22398 & 24940 & 14235 & 14848 & 5005 & 405 & 4548 & 12340\\
Qwen2.5-7B-Inst~+ Self-Retro-R1-7B & 21388 & 21175 & 10861 & 13843 & 5332 & 452 & 4302 & 11050\\
Qwen2.5-7B-Inst~+ R1-671B & 24540 & 27839 & 14750 & 15943 & 7726 & 2148 & 7170 & 14302\\
Qwen2.5-7B-Inst~+ W2S-Retro-R1-32B & 23284 & 25984 & 13418 & 16084 & 7480 & 2048 & 6687 & 13569\\
R1-distill Qwen2.5-7B~+ R1-7B & 20541 & 15381 & 7695 & 11012 & 3956 & 446 & 3807 & 8977\\
R1-distill Qwen2.5-7B~+ Self-Retro-R1-7B & 16965 & 13512 & 5683 & 8551 & 3043 & 451 & 2859 & 7295\\
R1-distill Qwen2.5-7B~+ R1-671B & 18140 & 16875 & 8299 & 11254 & 4867 & 1564 & 4927 & 9418\\
R1-distill Qwen2.5-7B~+ W2S-Retro-R1-32B & 16525 & 14636 & 8689 & 11113 & 4970 & 1665 & 4004 & 8800\\
Qwen2.5-32B-Instruct~+ R1-671B & 14311 & 11742 & 5983 & 9004 & 3474 & 1109 & 3893 & 7074\\
Qwen2.5-32B-Instruct~+ W2S-Retro-R1-32B & 12560 & 12385 & 6415 & 8280 & 3416 & 990 & 3618 & 6809\\
R1-distill Qwen2.5-32B~+ R1-671B (12k) & 12367 & 10798 & 5780 & 7817 & 3661 & 1183 & 3682 & 6470\\
R1-distill Qwen2.5-32B~+ W2S-Retro-R1-32B (12k) & 11575 & 11488 & 4918 & 7574 & 3136 & 932 & 3014 & 6091\\
\bottomrule
\end{tabular}
}\end{adjustbox}
\caption{Per-dataset evaluation results (response token length) using greedy decoding.}
\label{tab:appendix_finegrained_greedy_len}

\end{table}
\begin{table}[h!]
\centering
\tiny
\begin{adjustbox}{angle=270}{
\begin{tabular}{lllllllll}
\toprule
\textbf{Models (Temperature sampling, T=0.6 \& top-p=0.95)} & AIME25 & AIME24 & AMC23 & GaoKao23En & OlympiadBench & GSM8K & MATH500 & Avg. \\
\midrule
Qwen2.5-7B-Inst & 5.98 $\pm$ 1.74 & 12.68 $\pm$ 1.47 & 50.50 $\pm$ 1.48 & 38.90 $\pm$ 0.28 & 63.56 $\pm$ 0.76 & 88.76 $\pm$ 0.28 & 74.72 $\pm$ 0.36 & 47.87 $\pm$ 0.91\\
Qwen2.5-Math-7B & 6.68 $\pm$ 0.95 & 15.32 $\pm$ 2.02 & 51.00 $\pm$ 4.45 & 22.54 $\pm$ 1.93 & 45.46 $\pm$ 1.29 & 71.28 $\pm$ 0.97 & 60.72 $\pm$ 1.39 & 39.00 $\pm$ 1.86\\
Qwen2.5-Math-7B-Inst & 12.00 $\pm$ 1.52 & 14.66 $\pm$ 2.02 & 58.50 $\pm$ 2.61 & 39.42 $\pm$ 0.18 & 66.08 $\pm$ 0.41 & 95.36 $\pm$ 0.04 & 82.60 $\pm$ 0.42 & 52.66 $\pm$ 1.03\\
OpenR1-Qwen-7B & 40.68 $\pm$ 1.73 & 48.66 $\pm$ 2.22 & 84.00 $\pm$ 2.30 & 59.68 $\pm$ 0.63 & 81.56 $\pm$ 0.13 & 95.12 $\pm$ 0.18 & 92.08 $\pm$ 0.43 & 71.68 $\pm$ 1.09\\
R1-distill Qwen2.5-7B & 40.00 $\pm$ 1.32 & 55.32 $\pm$ 2.42 & 90.50 $\pm$ 0.84 & 57.80 $\pm$ 0.31 & 75.00 $\pm$ 0.73 & 86.72 $\pm$ 0.32 & 91.92 $\pm$ 0.39 & 71.04 $\pm$ 0.90\\
OpenThinker-7B & 28.00 $\pm$ 1.20 & 27.98 $\pm$ 1.79 & 68.50 $\pm$ 2.07 & 49.54 $\pm$ 0.28 & 72.92 $\pm$ 0.77 & 80.70 $\pm$ 0.55 & 85.72 $\pm$ 0.45 & 59.05 $\pm$ 1.02\\
Qwen2.5-32B-Instruct & 15.32 $\pm$ 1.79 & 14.02 $\pm$ 1.74 & 65.00 $\pm$ 1.58 & 46.98 $\pm$ 0.43 & 72.60 $\pm$ 0.19 & 95.44 $\pm$ 0.11 & 81.92 $\pm$ 0.54 & 55.90 $\pm$ 0.91\\
QwQ-32B-Preview & 34.02 $\pm$ 3.58 & 36.64 $\pm$ 1.89 & 82.50 $\pm$ 1.41 & 58.86 $\pm$ 0.42 & 80.20 $\pm$ 0.38 & 95.38 $\pm$ 0.20 & 90.24 $\pm$ 0.41 & 68.26 $\pm$ 1.18\\
Sky-T1-32B-Preview & 24.66 $\pm$ 0.74 & 27.32 $\pm$ 1.74 & 73.50 $\pm$ 0.89 & 54.28 $\pm$ 0.32 & 76.20 $\pm$ 0.55 & 95.86 $\pm$ 0.12 & 88.20 $\pm$ 0.22 & 62.86 $\pm$ 0.66\\
R1-distill Qwen2.5-32B & 57.34 $\pm$ 1.74 & 64.02 $\pm$ 3.32 & 95.00 $\pm$ 1.58 & 63.82 $\pm$ 0.47 & 78.90 $\pm$ 0.46 & 92.48 $\pm$ 0.20 & 92.60 $\pm$ 0.46 & 77.74 $\pm$ 1.18\\
OpenThinker-32B & 51.34 $\pm$ 1.78 & 61.36 $\pm$ 2.02 & 94.50 $\pm$ 0.84 & 64.64 $\pm$ 0.38 & 80.90 $\pm$ 0.33 & 85.46 $\pm$ 0.20 & 93.16 $\pm$ 0.27 & 75.91 $\pm$ 0.83\\
Qwen2.5-7B-Inst~+ R1-7B & 23.34 $\pm$ 2.11 & 22.68 $\pm$ 1.74 & 59.50 $\pm$ 2.17 & 46.92 $\pm$ 0.30 & 67.86 $\pm$ 0.78 & 86.68 $\pm$ 0.17 & 81.08 $\pm$ 0.43 & 55.44 $\pm$ 1.10\\
Qwen2.5-7B-Inst~+ R1-7B-shortest & 26.00 $\pm$ 1.98 & 26.66 $\pm$ 2.66 & 62.00 $\pm$ 1.64 & 42.58 $\pm$ 0.57 & 62.82 $\pm$ 0.45 & 85.50 $\pm$ 0.41 & 76.40 $\pm$ 0.27 & 54.57 $\pm$ 1.14\\
Qwen2.5-7B-Inst~+ Self-Retro-R1-7B & 25.34 $\pm$ 2.02 & 26.70 $\pm$ 0.00 & 61.00 $\pm$ 2.88 & 44.52 $\pm$ 0.30 & 67.08 $\pm$ 0.60 & 86.70 $\pm$ 0.19 & 79.56 $\pm$ 0.56 & 55.84 $\pm$ 0.94\\
Qwen2.5-7B-Inst~+ R1-671B & 24.66 $\pm$ 0.74 & 30.00 $\pm$ 2.83 & 71.00 $\pm$ 1.52 & 49.36 $\pm$ 0.40 & 71.84 $\pm$ 0.39 & 77.60 $\pm$ 0.24 & 84.52 $\pm$ 0.44 & 58.43 $\pm$ 0.94\\
Qwen2.5-7B-Inst~+ W2S-Retro-R1-32B & 20.68 $\pm$ 1.46 & 24.64 $\pm$ 1.20 & 67.50 $\pm$ 3.00 & 47.74 $\pm$ 0.43 & 72.16 $\pm$ 0.46 & 87.62 $\pm$ 0.30 & 84.20 $\pm$ 0.51 & 57.79 $\pm$ 1.05\\
R1-distill Qwen2.5-7B~+ R1-7B & 38.00 $\pm$ 2.77 & 60.02 $\pm$ 2.30 & 90.00 $\pm$ 1.22 & 59.84 $\pm$ 0.55 & 75.06 $\pm$ 0.25 & 85.42 $\pm$ 0.22 & 90.52 $\pm$ 0.26 & 71.27 $\pm$ 1.08\\
R1-distill Qwen2.5-7B~+ Self-Retro-R1-7B & 44.00 $\pm$ 4.46 & 51.34 $\pm$ 4.38 & 89.50 $\pm$ 1.30 & 59.30 $\pm$ 0.48 & 74.74 $\pm$ 0.85 & 84.76 $\pm$ 0.23 & 90.60 $\pm$ 0.25 & 70.61 $\pm$ 1.71\\
R1-distill Qwen2.5-7B~+ R1-671B & 41.32 $\pm$ 2.92 & 54.02 $\pm$ 2.19 & 92.00 $\pm$ 1.30 & 60.90 $\pm$ 0.51 & 80.16 $\pm$ 0.50 & 82.00 $\pm$ 0.32 & 91.36 $\pm$ 0.24 & 71.68 $\pm$ 1.14\\
R1-distill Qwen2.5-7B~+ W2S-Retro-R1-32B & 40.00 $\pm$ 1.32 & 58.00 $\pm$ 2.22 & 90.50 $\pm$ 1.48 & 60.46 $\pm$ 0.28 & 81.52 $\pm$ 0.45 & 89.56 $\pm$ 0.25 & 91.56 $\pm$ 0.24 & 73.09 $\pm$ 0.89\\
Qwen2.5-32B-Instruct~+ R1-671B & 50.00 $\pm$ 3.77 & 59.98 $\pm$ 2.30 & 94.50 $\pm$ 1.30 & 65.02 $\pm$ 0.43 & 81.24 $\pm$ 0.38 & 85.44 $\pm$ 0.27 & 93.08 $\pm$ 0.27 & 75.61 $\pm$ 1.25\\
Qwen2.5-32B-Instruct~+ W2S-Retro-R1-32B & 50.00 $\pm$ 2.11 & 67.34 $\pm$ 1.97 & 92.50 $\pm$ 1.58 & 63.30 $\pm$ 0.32 & 83.30 $\pm$ 0.25 & 92.60 $\pm$ 0.19 & 93.08 $\pm$ 0.37 & 77.45 $\pm$ 0.97\\
R1-distill Qwen2.5-32B~+ R1-671B (12k) & 63.34 $\pm$ 2.49 & 74.66 $\pm$ 2.02 & 94.00 $\pm$ 0.89 & 67.94 $\pm$ 0.31 & 82.04 $\pm$ 0.25 & 83.00 $\pm$ 0.31 & 93.92 $\pm$ 0.28 & 79.84 $\pm$ 0.94\\
R1-distill Qwen2.5-32B~+ W2S-Retro-R1-32B (12k)  & 60.02 $\pm$ 2.50 & 70.02 $\pm$ 2.30 & 97.50 $\pm$ 1.22 & 66.50 $\pm$ 0.45 & 84.16 $\pm$ 0.29 & 94.64 $\pm$ 0.10 & 94.40 $\pm$ 0.25 & 81.03 $\pm$ 1.02\\
\bottomrule
\end{tabular}
}\end{adjustbox}
\caption{Per-dataset evaluation results (accuracies) using temperature sampling (t=0.6 and top-p=0.95). The numbers after $\pm$ means the 95\% confidence interval.}
\label{tab:appendix_finegrained_temp_acc}

\end{table}

\begin{table}[h!]
\centering
\tiny
\begin{adjustbox}{angle=270}{
\begin{tabular}{lllllllll}
\toprule
\textbf{Models (Temperature sampling, T=0.6 \& top-p=0.95)} & AIME25 & AIME24 & AMC23 & GaoKao23En & OlympiadBench & GSM8K & MATH500 & Avg. \\
\midrule
Qwen2.5-7B-Inst & 888 $\pm$ 25 & 1628 $\pm$ 399 & 2051 $\pm$ 348 & 1116 $\pm$ 42 & 653 $\pm$ 41 & 258 $\pm$ 2 & 637 $\pm$ 44 & 1033 $\pm$ 129\\
Qwen2.5-Math-7B & 1545 $\pm$ 119 & 1492 $\pm$ 139 & 1148 $\pm$ 92 & 1602 $\pm$ 86 & 1071 $\pm$ 38 & 681 $\pm$ 23 & 1038 $\pm$ 40 & 1225 $\pm$ 77\\
Qwen2.5-Math-7B-Inst & 1451 $\pm$ 89 & 1434 $\pm$ 142 & 1091 $\pm$ 73 & 1124 $\pm$ 11 & 779 $\pm$ 16 & 330 $\pm$ 2 & 686 $\pm$ 8 & 985 $\pm$ 49\\
OpenR1-Qwen-7B & 13960 $\pm$ 228 & 14344 $\pm$ 302 & 7421 $\pm$ 212 & 9100 $\pm$ 66 & 4330 $\pm$ 33 & 1392 $\pm$ 6 & 3635 $\pm$ 27 & 7740 $\pm$ 125\\
R1-distill Qwen2.5-7B & 13808 $\pm$ 322 & 12642 $\pm$ 318 & 5785 $\pm$ 112 & 8502 $\pm$ 72 & 3163 $\pm$ 106 & 446 $\pm$ 6 & 3470 $\pm$ 53 & 6831 $\pm$ 141\\
OpenThinker-7B & 17009 $\pm$ 795 & 17750 $\pm$ 432 & 11227 $\pm$ 198 & 10866 $\pm$ 113 & 5181 $\pm$ 115 & 1404 $\pm$ 30 & 5409 $\pm$ 63 & 9835 $\pm$ 250\\
Qwen2.5-32B-Instruct & 1080 $\pm$ 184 & 1349 $\pm$ 305 & 691 $\pm$ 9 & 946 $\pm$ 50 & 519 $\pm$ 13 & 229 $\pm$ 2 & 511 $\pm$ 13 & 761 $\pm$ 82\\
QwQ-32B-Preview & 9589 $\pm$ 252 & 10109 $\pm$ 543 & 4601 $\pm$ 407 & 5827 $\pm$ 46 & 2578 $\pm$ 61 & 757 $\pm$ 17 & 2683 $\pm$ 53 & 5163 $\pm$ 197\\
Sky-T1-32B-Preview & 1899 $\pm$ 179 & 4123 $\pm$ 523 & 3273 $\pm$ 566 & 2382 $\pm$ 72 & 989 $\pm$ 55 & 300 $\pm$ 15 & 1162 $\pm$ 26 & 2018 $\pm$ 205\\
R1-distill Qwen2.5-32B & 12644 $\pm$ 548 & 11258 $\pm$ 501 & 5639 $\pm$ 270 & 7269 $\pm$ 45 & 2971 $\pm$ 100 & 444 $\pm$ 2 & 2984 $\pm$ 97 & 6173 $\pm$ 223\\
OpenThinker-32B & 13474 $\pm$ 365 & 11885 $\pm$ 403 & 6161 $\pm$ 175 & 7960 $\pm$ 56 & 3600 $\pm$ 45 & 1102 $\pm$ 6 & 3696 $\pm$ 53 & 6840 $\pm$ 157\\
Qwen2.5-7B-Inst~+ R1-7B & 15783 $\pm$ 330 & 17915 $\pm$ 757 & 9586 $\pm$ 66 & 10612 $\pm$ 97 & 4112 $\pm$ 70 & 456 $\pm$ 1 & 4253 $\pm$ 114 & 8959 $\pm$ 205\\
Qwen2.5-7B-Inst~+ R1-7B-shortest & 14468 $\pm$ 948 & 17526 $\pm$ 297 & 8691 $\pm$ 494 & 8836 $\pm$ 69 & 3007 $\pm$ 85 & 398 $\pm$ 4 & 3138 $\pm$ 55 & 8009 $\pm$ 279\\
Qwen2.5-7B-Inst~+ Self-Retro-R1-7B & 15211 $\pm$ 601 & 17547 $\pm$ 820 & 8039 $\pm$ 465 & 9485 $\pm$ 69 & 3623 $\pm$ 93 & 455 $\pm$ 3 & 3483 $\pm$ 123 & 8263 $\pm$ 310\\
Qwen2.5-7B-Inst~+ R1-671B & 17144 $\pm$ 279 & 18529 $\pm$ 648 & 9942 $\pm$ 338 & 11139 $\pm$ 131 & 5216 $\pm$ 96 & 1441 $\pm$ 15 & 5359 $\pm$ 37 & 9824 $\pm$ 220\\
Qwen2.5-7B-Inst~+ W2S-Retro-R1-32B & 15287 $\pm$ 404 & 16964 $\pm$ 318 & 9719 $\pm$ 501 & 10062 $\pm$ 65 & 4561 $\pm$ 62 & 1170 $\pm$ 9 & 4817 $\pm$ 71 & 8940 $\pm$ 204\\
R1-distill Qwen2.5-7B~+ R1-7B & 12727 $\pm$ 419 & 11385 $\pm$ 384 & 5906 $\pm$ 171 & 7808 $\pm$ 70 & 2860 $\pm$ 102 & 456 $\pm$ 5 & 3243 $\pm$ 76 & 6341 $\pm$ 175\\
R1-distill Qwen2.5-7B~+ Self-Retro-R1-7B & 11506 $\pm$ 317 & 10119 $\pm$ 394 & 4316 $\pm$ 269 & 6529 $\pm$ 71 & 2343 $\pm$ 23 & 451 $\pm$ 2 & 2575 $\pm$ 63 & 5406 $\pm$ 162\\
R1-distill Qwen2.5-7B~+ R1-671B & 13730 $\pm$ 379 & 12516 $\pm$ 483 & 6309 $\pm$ 151 & 8632 $\pm$ 65 & 3734 $\pm$ 63 & 1337 $\pm$ 15 & 3944 $\pm$ 49 & 7172 $\pm$ 172\\
R1-distill Qwen2.5-7B~+ W2S-Retro-R1-32B & 13594 $\pm$ 703 & 11087 $\pm$ 344 & 5409 $\pm$ 242 & 7810 $\pm$ 86 & 3273 $\pm$ 33 & 1144 $\pm$ 22 & 3429 $\pm$ 26 & 6535 $\pm$ 208\\
Qwen2.5-32B-Instruct~+ R1-671B & 12765 $\pm$ 407 & 11778 $\pm$ 262 & 6080 $\pm$ 181 & 7867 $\pm$ 35 & 3478 $\pm$ 15 & 1117 $\pm$ 6 & 3649 $\pm$ 61 & 6676 $\pm$ 138\\
Qwen2.5-32B-Instruct~+ W2S-Retro-R1-32B & 11935 $\pm$ 367 & 9866 $\pm$ 199 & 5434 $\pm$ 88 & 7010 $\pm$ 25 & 3132 $\pm$ 27 & 927 $\pm$ 6 & 3154 $\pm$ 22 & 5923 $\pm$ 105\\
R1-distill Qwen2.5-32B~+ R1-671B (12k)  & 12075 $\pm$ 268 & 9802 $\pm$ 282 & 5827 $\pm$ 171 & 7442 $\pm$ 47 & 3382 $\pm$ 28 & 1165 $\pm$ 5 & 3460 $\pm$ 27 & 6164 $\pm$ 118\\
R1-distill Qwen2.5-32B~+ W2S-Retro-R1-32B (12k)& 10390 $\pm$ 179 & 9078 $\pm$ 384 & 4463 $\pm$ 113 & 6481 $\pm$ 38 & 2882 $\pm$ 32 & 905 $\pm$ 13 & 2907 $\pm$ 11 & 5301 $\pm$ 110\\
\bottomrule
\end{tabular}
}\end{adjustbox}
\caption{Per-dataset evaluation results (model response token length) using temperature sampling (t=0.6 and top-p=0.95). The numbers after $\pm$ means the 95\% confidence interval.}
\label{tab:appendix_finegrained_temp_len}

\end{table}
In Tables~\ref{tab:appendix_finegrained_greedy_acc} and \ref{tab:appendix_finegrained_greedy_len}, we share the per-dataset evaluation results using greedy decoding, and in Tables~\ref{tab:appendix_finegrained_temp_acc} and \ref{tab:appendix_finegrained_temp_len}, we share results using temperature sampling with top-p=0.95 and T=0.6. We use the max response length of 32,768 tokens for all experiments. For temperature sampling, we use random five seeds and aggregate the results, and we further report the confidence interval to share the deviation of the metrics.

\end{appendices}

\clearpage

\end{document}